\pdfoutput=1
\relax
\documentclass[letterpaper]{article} % DO NOT CHANGE THIS
\usepackage{aaai22}  % DO NOT CHANGE THIS
\usepackage{times}  % DO NOT CHANGE THIS
\usepackage{helvet}  % DO NOT CHANGE THIS
\usepackage{courier}  % DO NOT CHANGE THIS
\usepackage[hyphens]{url}  % DO NOT CHANGE THIS
\usepackage{graphicx} % DO NOT CHANGE THIS
\usepackage{natbib}  % DO NOT CHANGE THIS AND DO NOT ADD ANY OPTIONS TO IT
\usepackage{caption} % DO NOT CHANGE THIS AND DO NOT ADD ANY OPTIONS TO IT
\DeclareCaptionStyle{ruled}{labelfont=normalfont,labelsep=colon,strut=off} % DO NOT CHANGE THIS
\usepackage{algorithm}
\usepackage{algorithmic}

\usepackage{amsmath}
\usepackage{dsfont}
\usepackage{multirow}
\usepackage{bm}

\usepackage{newtxmath}

\usepackage{array}
\usepackage{newfloat}
\usepackage{listings}

\usepackage{color,xcolor}

\floatstyle{ruled}
\newfloat{listing}{tb}{lst}{}
\floatname{listing}{Listing}
\title{When Facial Expression Recognition Meets Few-Shot Learning: A Joint and Alternate Learning Framework}
\author{
   Xinyi Zou\textsuperscript{\rm 1},
   ~~
   Yan Yan\textsuperscript{\rm 1}\thanks{Corresponding author (email: {\tt yanyan@xmu.edu.cn}).},
   ~~
   Jing-Hao Xue\textsuperscript{\rm 2},
   ~~
   Si Chen\textsuperscript{\rm 3},
   ~~
   Hanzi Wang\textsuperscript{\rm 1}\\

}

\affiliations{
	    \textsuperscript{\rm 1}Xiamen University, China
	~~~~\textsuperscript{\rm 2}University College London, UK
	~~~~
	\textsuperscript{\rm 3}Xiamen University of Technology, China
	
}

\usepackage{bibentry}
\begin{document}

\maketitle

\begin{abstract}
Human emotions involve basic and compound facial expressions. However, current research on facial expression recognition (FER) mainly focuses on basic expressions, and thus fails to address the diversity of human emotions in practical scenarios. Meanwhile, existing work on compound FER relies heavily on abundant labeled compound expression training data, which are often laboriously collected under the professional instruction of psychology. In this paper, we study compound FER in the cross-domain few-shot learning setting, where only a few images of novel classes from the target domain are required as a reference. In particular, we aim to identify unseen compound expressions with the model trained on easily accessible basic expression datasets. To alleviate the problem of limited base classes in our FER task, we propose a novel Emotion Guided Similarity Network (EGS-Net), consisting of an emotion branch and a similarity branch, based on a two-stage learning framework. Specifically, in the first stage, the similarity branch is jointly trained with the emotion branch in a multi-task fashion. With the regularization of the emotion branch, we prevent the similarity branch from overfitting to sampled base classes that are highly overlapped across different episodes. In the second stage, the emotion branch and the similarity branch play a ``two-student game'' to alternately learn from each other, thereby further improving the inference ability of the similarity branch on unseen compound expressions. Experimental results on both in-the-lab and in-the-wild compound expression datasets demonstrate the superiority of our proposed method against several state-of-the-art methods.
\end{abstract}

\section{Introduction}

%\begin{figure}[!t]
%	\centering
%	\includegraphics[scale=0.55]{figures/motivation.pdf}
%	\caption{Given only a few reference images, humans can easily recognize an \textcolor{red}{unseen expression} based on the prior knowledge of \textcolor{red}{various seen} expressions.}
%	\label{fig:motivation}
%\end{figure}

%As one of the most natural and important ways for humans to express their emotions and intentions, facial expression is of great significance in interpersonal communication.
Over the past few decades, facial expression recognition (FER) has attracted considerable attention because of its wide range of applications in {human-robot} interaction, online education, driver monitoring, \emph{etc} \cite{corneanu2016survey}.

Based on Ekman and Friesen's study \cite{ekman1971constants}, facial expressions are typically classified into seven basic expressions, including happiness, sadness, disgust, anger, fear, surprise, and neutral.
Previous work on FER chiefly focuses on the classification of these pre-defined basic expressions.  Accordingly, numerous basic expression datasets \cite{lucey2010extended,li2017reliable,zhao2011facial} have been collected, and impressive progress \cite{li2018occlusion, ruan2020deep,zhao2021robust} has been made to address large facial appearance variations caused by identity, pose, occlusion, illumination, and so on.
In this paper, we refer to the above conventional FER task as the basic FER task.

Regrettably, these basic expressions cannot completely characterize the diversity of human emotions in nature.
Du \emph{et al.} \cite{du2014compound} reveal that human emotions involve compound expressions, beyond the above basic expressions. They enlarge the number of expressions to 22 by combining basic expressions.
%EmotioNet and icv-MEFED \cite{guo2018dominant} datasets further study compound expressions by using an AU-based annotation mechanism and defining dominant-complementary expressions, respectively.
Later, the EmotioNet dataset \cite{fabian2016emotionet} is constructed with large-scale compound expression data.
%annotates compound expressions by using an AU-based annotation mechanism.
%\textcolor{blue}{The icv-MEFED dataset \cite{guo2018dominant} further introduces more compound expressions by defining dominant-complementary expressions.(delete?)}
%\textcolor{red}
{To classify the above compound expressions, conventional deep learning based methods \cite{slimani2019compound, guo2017multi} usually rely heavily on a large amount of labeled compound expression training data. However, collecting such data is laborious and often demands the professional instruction of psychology.} % \textcolor{blue}{In \cite{liang2020fine, wang2019fine}, fine-grained expressions are annotated to explore the complexity of human emotions. (delete?)}

%\textcolor{blue}{Note that it is not a trivial task to define all possible expression categories due to the great uncontrollability and variety of human emotions.}

 %\textcolor{red}{Therefore, a much flexible paradigm for FER task needs to be discovered.}

As humans, given only a few reference images (a support set), we can easily recognize an {unseen expression} (a query) based on the prior knowledge of {various seen} expressions. Recent research on few-shot learning (FSL) exhibits the potential of quickly generalizing to novel classes with only a few labeled data of these classes, thereby reducing the gap between humans and artificial intelligence \cite{lu2020learning}.
%identifying unseen classes without requiring a great number of newly labeled training data.
In this paper, we investigate compound FER in the cross-domain FSL (CD-FSL) paradigm, which greatly alleviates the burden of collecting  large-scale labeled compound expression data.
%and predefining all possible expression categories}.
%, which identifies unseen compound expressions by using the model only trained on basic expressions.
Notably, instead of manually splitting a compound expression dataset into a base class set and a novel class set, we explore a more challenging but practical setup,
which aims to classify compound expressions from the unseen domain by using the model  trained only on easily accessible basic expression datasets.
%where the easily accessible basic FER datasets are used for training and the compound FER datasets are used for testing.

Traditional FSL methods have achieved promising performance in many computer vision tasks, such as image classification \cite{li2019distribution,yao2020graph} and object detection \cite{dong2018few, yang2020context}. However, few work is concerned with the compound FER task in the CD-FSL setting. Different from widely used FSL benchmarks (e.g., miniImageNet \cite{vinyals2016matching} and Ominiglot \cite{lake2015human} whose total numbers of classes are 100 and 1,623, respectively), basic expression datasets contain a limited number of basic expressions (i.e., base classes in our setting).
Consequently, the random sampling process cannot effectively simulate the variance of unseen tasks since the sampled base classes are highly overlapped across different episodes. In this way, traditional FSL methods easily suffer from the overfitting problem, resulting in their deteriorated inference ability on unseen compound expressions.  
%and thus the  is substantially affected.
%\textcolor{blue}{Therefore, the random sampling process from base classes cannot effectively simulate the variance of unseen tasks, which leads to the overfitting problem of the highly-overlapped sampled classes. The inference ability of these methods on unseen compound expressions is far from satisfactory.}

%\textcolor{red}{Therefore, the random sampling classes highly overlap between , and the learned model easily suffers from the overfitting problem. As a result, }

%As a result, the inference ability of traditional FSL methods is substantially affected and the performance of these methods is still far from being satisfactory for compound FER.

To address the above problem, we propose an effective CD-FSL method called {Emotion Guided} Similarity Network (EGS-Net), consisting of an emotion branch and a similarity branch, for compound FER. The emotion branch captures the global information of basic expressions and serves as a regularizer, while the similarity branch learns a transferable similarity metric between two expressions.
%for unseen compound expressions.
%a\textcolor{red}{Considering the difficulty of directly improve the inference ability on the unseen task}
In particular, motivated by the human perception that one can better identify compound expressions with more prior knowledge of basic expressions, we develop a two-stage learning framework to train EGS-Net in a progressive manner: (i) joint learning of the emotion branch and the similarity branch in a multi-task fashion;
%, to improve the generalization ability on basic expressions from the unseen domain; %mitigate the model from overfitting on the \textcolor{red}{base} sampled classes;
(ii) alternate learning between the emotion branch and the similarity branch.
%, to further improve the inference ability on unseen compound expressions.
As a result, our proposed method remarkably relieves the requirement of abundant compound expression training data and offers superior scalability for practical applications.

%To be specific, in the first stage, we jointly train the emotion branch and the similarity branch to exploit the knowledge of basic expressions. In this stage, the emotion branch shares the global information of basic expressions with the similarity branch, avoiding the similarity branch being trapped into highly overlapped sampled base classes.
%Hence, the inference ability of the similarity branch on basic expressions from the unseen domain is significantly strengthened. However, its performance on unseen compound expressions is still inferior due to the poor inference ability of the initial emotion branch on novel classes.
%inferior due to the \textcolor{blue}{regularization of the emotion branch} in this stage.
%\textcolor{red}{Though better generalized to the unseen domain on seen classes, the performance on the unseen classes is still limited due to the inferior inference ability of the conventional emotion branch.}
%In the second stage, the emotion branch and the similarity branch play a ``two-student game'' by alternate learning, where one branch learns from the other one in turn. Instead of competing with each other in an adversarial process as generative adversarial network (GAN) \cite{goodfellow2014generative}, alternate learning takes advantage of the two students to improve each other's performance from a different perspective. Such a manner enhances the inference ability of our model, especially on unseen compound expressions.

In summary, our main contributions are given as follows:
\begin{itemize}
	\item We propose a novel EGS-Net method for compound FER in the CD-FSL setting.
	Our method is capable of learning a transferable model, which is trained only on multiple basic expression datasets.
	Therefore, we can easily recognize novel compound expressions from the unseen domain, with a few reference images of novel classes. To the best of our knowledge, we are the first to classify unseen compound expressions in the FSL scenario.
	
	\item We develop a two-stage learning framework to progressively train EGS-Net and thus effectively alleviate the problem of limited base classes in our FER task.
	Based on the proposed learning framework, the inference ability of the similarity branch can be greatly improved with the help of the emotion branch, thereby boosting the performance of predicting novel compound expressions.
	
	\item Extensive experimental results on both in-the-lab and in-the-wild compound expression datasets demonstrate the effectiveness of our proposed method in comparison with several state-of-the-art FSL methods. %We further extend our method to representative learn-to-measure baselines to show its potential extensibility.
\end{itemize}

\section{Related Work}

%In this section, we
%give a brief review of the related work
%from three aspects, i.e.,
%basic facial expression recognition, compound facial expression recognition, and few-shot learning.

%\subsection{Basic Facial Expression Recognition}

\noindent \textbf{Facial Expression Recognition.}
The past decades have witnessed significant progress in FER. Considering its practical applications, the main focus of FER has shifted from controllable in-the-lab scenarios to more challenging in-the-wild scenarios. However, conventional FER methods  \cite{li2018occlusion, ruan2020deep, zhao2021robust} only classify basic expressions, and fail to depict the complexity of human emotions in practical scenarios.
%\textcolor{blue}{, where facial images often suffer from occlusions, large pose variations, {extreme illumination changes}, \emph{etc}.}

%Many methods \cite{wang2020region, ruan2020deep} leverage the attention mechanisms or disturbance-disentangling to address these disturbances.
%\textcolor{red}{The attention-based and disentangling-based methods are most commonly used to handle the disturbance \cite{wang2020region, ruan2020deep}.}

%\textcolor{blue}{Wang \emph{et al.} \cite{wang2020region} propose a novel region attention network to address the occlusion and pose variation problems for basic FER, while Ruan \emph{et al.} \cite{ruan2020deep} introduce a deep disturbance-disentangled learning method to explicitly disentangle multiple disturbing factors.}%

%Li \emph{et al.} %\cite{li2018occlusion} develop an attention based network to perceive occluded regions and force the model to focus on non-occluded regions. Wang \emph{et al.} \cite{wang2019identity} perform identity- and pose-robust basic expression recognition based on adversarial feature learning. Recently, Ruan \emph{et al.} \cite{ruan2020deep} introduce a deep disturbance-disentangled learning method to explicitly disentangle multiple disturbing factors and achieve promising {results} on basic FER datasets.

Recently, Du \emph{et al.} \cite{du2014compound} reveal that there are a large number of emotions expressed regularly by humans. They further define the compound expressions by combining basic expressions.
%A computational model is used to show the distinct differences among these compound expressions.
%Though basic expression is well studied, detailed facial expression recognition received less attention due to the limitation of the dataset.  The pioneer work of compound expression is \cite{du2014compound}, which first defined the compound emotion categories by combining two basic emotions.
Benitez-Quiroz \emph{et al.} \cite{fabian2016emotionet} introduce a large compound expression dataset called EmotioNet, which contains one million in-the-wild images labeled by an AU-based algorithm. %Later, Guo \emph{et al.} \cite{guo2018dominant} release the icv-MEFED dataset, which is collected under the careful supervision of psychologists. They increase the number of compound expressions to 50 by treating symmetric categories  (e.g., surprisingly-happy and happily-surprised) as different classes.
%To further study the diversity of facial expressions, several fine-grained expression datasets \cite{liang2020fine, wang2019fine} are also collected.
Based on the above datasets, several attempts are made for compound FER. Slimani \emph{et al.} \cite{slimani2019compound} propose a highway convolutional neural network which replaces the shortcut with a learnable parameter for compound FER.
As a winner of the FG 2017 Challenge, Guo \emph{et al.} \cite{guo2017multi} design a multi-modality convolutional neural network, which combines the visual feature with the geometry feature and shows superiority for the emotion challenge.

{Conventional compound FER methods require a large amount of labeled compound expression training data.} Collecting such data not only is time-c7onsuming and labor-intensive, but also demands the professional guidance. %Moreover, it is a challenging task to pre-define all possible expression categories.
In this paper, different from the above methods, we are concerned with the compound FER problem in the CD-FSL setting,
where the base classes are sampled from multiple basic expression datasets and the novel classes are  compound expressions.
%Therefore, we manage to greatly alleviate the requirement of using a large-scale labeled compound FER dataset for training, and are able to easily classify novel compound expressions with only a few labeled reference images.
Therefore, we manage to perform compound FER with only a few labeled reference images and provide great flexibility to identify a new expression category.

%\subsection{Few-Shot Learning}
\noindent \textbf{Few-Shot Learning.}
%FSL aims at learning new concepts from extremely few samples.
%Early FSL methods \cite{miller2000learning, lake2015human} are mainly based on generative models.
With the success of convolutional neural networks, deep {learning based} FSL methods have become topical. 
These methods can be coarsely classified into meta-learning based methods \cite{vinyals2016matching, snell2017prototypical, sung2018learning, garcia2017few, finn2017model} and transfer learning based methods \cite{chen2019closer,afrasiyabi2020associative,hu2020leveraging,yang2021free}. In this paper, our method belongs to meta-learning based methods and it is based on the learn-to-measure (L2M) technique that aims to learn a transferable similarity metric.

Recently, some FSL methods \cite{luo2017label,tseng2020cross} are also developed under the cross-domain setting. %where the base classes and novel classes are sampled from different domains.
For example, Luo \emph{et al.} \cite{luo2017label} adopt adversarial learning to learn a transferable representation across different domains. Tseng \emph{et al.} \cite{tseng2020cross} propose novel feature-wise transformation layers to simulate the variance of the target domain. %, we adopt multiple basic FER datasets as multiple source domains to alleviate the domain gap between the training set and the test set. }
Guo \emph{et al.} \cite{guo2020broader} investigate a more challenging scenario, where a large domain shift exists between the base class domain and the novel class one.

Although existing FSL methods have shown promising performance in a variety of {computer vision tasks,} few of them study the compound FER task. {The most relevant work to ours is \cite{ciubotaru2019revisiting}, which evaluates some representative FSL methods for the basic FER task rather than generalizing to classify unseen compound expressions.} In fact, due to the limited number of base classes in our FER task, the performance of existing FSL methods drops substantially. Hence, we develop a novel Emotion Guided Similarity Network (EGS-Net) based on a two-stage learning framework to address this issue.

\begin{figure*}[th!]
	\centering
	\includegraphics[scale=0.75]{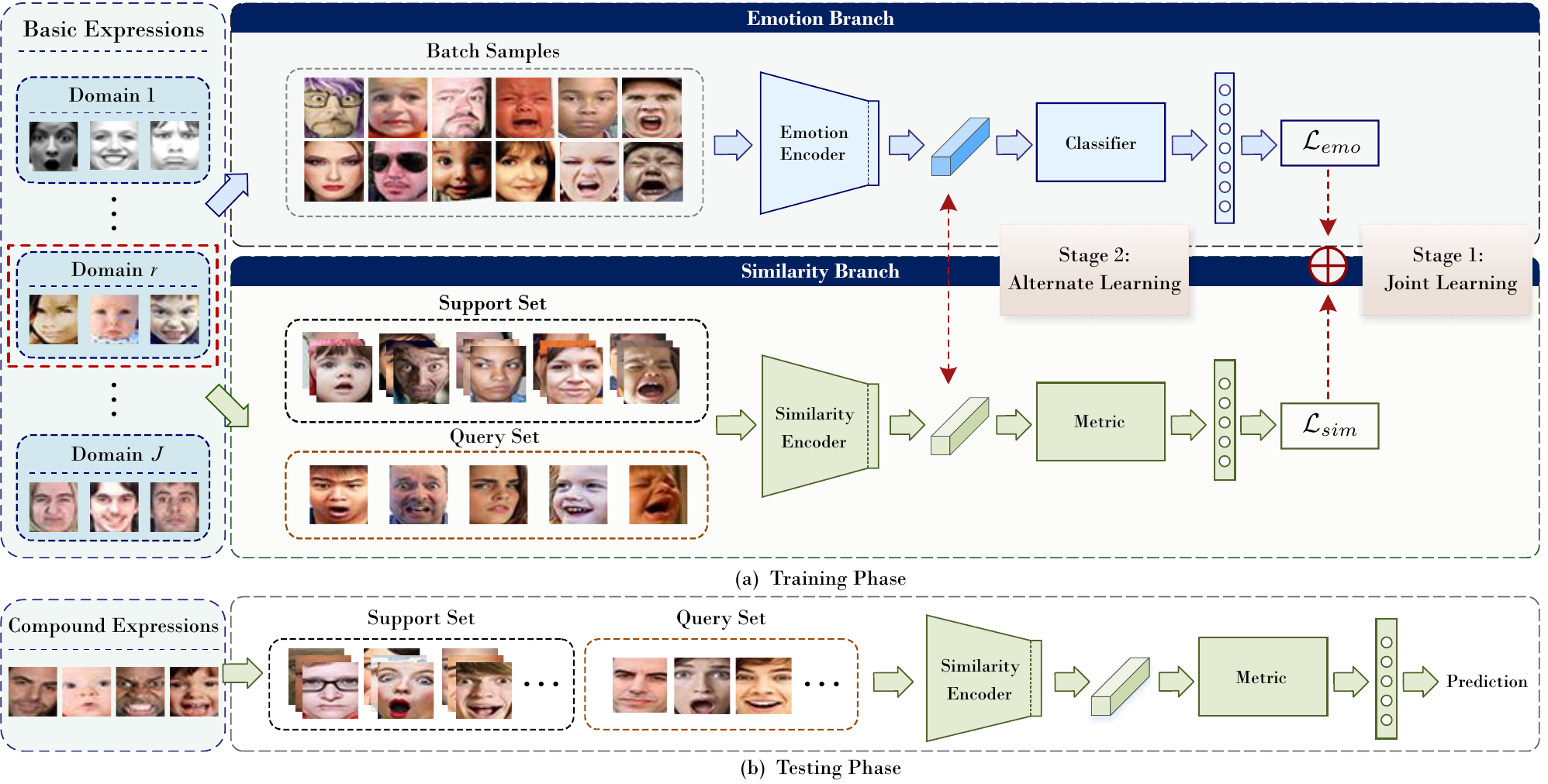}
	\caption{Overview of the proposed EGS-Net, which consists of an emotion branch and a similarity branch. (a) During the training phase, EGS-Net is progressively trained by using a two-stage learning framework. %textcolor{blue}{Multiple basic FER datasets are used as the training data.}
		In stage 1, we perform joint learning of the two branches in a multi-task fashion. In stage 2, we perform alternate learning between the two branches. (b) During the testing phase, the performance is evaluated on the compound expression dataset based on the learned similarity branch.}
	\label{fig:overview}
\end{figure*}

\section{Proposed Method}

%In this section, we introduce our proposed EGS-Net and {the} two-stage learning framework in detail. We first {give} the problem definition. Then, we {present} an overview of the proposed method. Next, we illustrate the details of {the developed} two-stage learning framework. Finally, we summarize the overall training {process}.

\subsection{Problem Definition}

In this paper, we perform compound FER in the CD-FSL setting, where the classes of the training set (i.e., the base class set) and those of the test set (i.e., the novel class set) are disjoint, and they are from different domains. %Specifically, we use the easily accessible basic FER datasets as the training set, and the compound FER datasets as the test set.
To enrich the diversity of base classes and bridge the domain gap between the training set and the test set, we adopt multiple source domains (i.e., multiple basic expression datasets) for training.
Accordingly, basic expressions from  source domains are introduced to construct the base classes and compound expressions from the target domain (i.e., a compound expression dataset) are used as the novel classes.
%enrich the diversity of base classes and
%to enrich the diversity of base classes and bridge the domain gap between the training set and the test set, %multiple source domains (i.e., multiple basic FER datasets)
%Basic expressions from multiple source domains are introduced to construct the base classes and compound expressions from the compound FER dataset are used as the novel classes.
Such a setting is a {challenging but practical} setup, which investigates the ability of recognizing novel compound expressions based on the model  trained only on easily accessible basic expression datasets. Therefore, given a base class set with sufficient labeled images, we aim to learn a transferable model and evaluate its performance on a novel class set with a few reference images. This enables the flexibility of the model to address compound FER.
%This greatly relieves the burden of collecting large-scale labeled compound FER datasets.
%and defining all possible expression categories in advance, enabling the flexibility of the model to address the compound FER task.

%Our method adopts the L2M technique, which manages to learn a transferable similarity metric by constructing similar tasks in two phases (meta-training and meta-testing). During the meta-training phase, a transferable model is trained in an episodic manner. In each episode, a meta-task is performed by selecting a support set and a small query set from the training set, and the model parameters are updated by the classification errors on the selected query set. During the meta-testing phase, based on the learned model, only a few reference images of novel classes are required to identify unseen categories. Such a manner enables the flexibility of the model to address the compound FER task.

\subsection{Overview}

%Different from conventional FSL benchmarks that contain a large number of classes,
%basic FER datasets involve only a few basic expressions (i.e., the base classes in our setting). The problem of limited base classes seriously affects the inference ability of current FSL methods. To address this, we propose a novel Emotion-Guided Similarity Network (EGS-Net) based on a two-stage learning framework, which effectively improves the inference ability of the model to identify novel compound expressions.

An overview of the proposed {Emotion Guided} Similarity Network (EGS-Net) is shown in Figure \ref{fig:overview}. EGS-Net consists of an emotion branch and a similarity branch. The emotion branch learns  global feature representations to classify all the basic expressions, while
the similarity branch learns a transferable similarity metric between two expressions. Specifically, for the training phase, the emotion branch is learned by mini-batch training. Meanwhile, the similarity branch follows the L2M setting and it is trained in an episodic manner.
%aims to learn} a transferable similarity metric that can be easily adapted to an unseen task. Moreover, multiple source domains are introduced to construct the training set, where a source domain is randomly chosen during every episode.
%The similarity branch learns a similarity metric between two expressions. In this paper, the similarity branch follows the L2M setting, which
%constructs similar tasks in an episodic meta-training manner.
%in two phases (meta-training and meta-testing). During the meta-training phase, a transferable model is trained in an episodic manner.
In each episode, a meta-task is performed by {sampling} a support set and a query set from a randomly selected source domain, and then the model parameters are updated by the classification errors on the {sampled} query set.  For the testing phase, we construct similar meta-tasks from the compound expression dataset.
In each meta-task, based on the learned similarity branch, a query image is classified into its nearest category in the support set.
%Based on the learned similarity branch,
%\textcolor{blue}{Based on the learned similarity branch, only a few reference images of \textcolor{red}{novel classes} are required to identify unseen compound expressions. (delete?)}

In particular, considering the difficulty of performing compound FER in the CD-FSL setting due to the
limited base classes, a two-stage learning framework (including a joint learning stage and an alternate learning stage) is developed to train EGS-Net progressively. In the first stage, the emotion branch and the similarity branch are jointly trained in a multi-task fashion. %In this stage, the emotion branch, which captures the global information of all the basic expressions, acts as a regularizer to avoid the similarity branch being trapped into highly-overlapped sampled base classes. %The inference ability of similarity branch on basic expressions from the unseen domain is substantially improved but that on compound expressions is still limited.
In the second stage, the two branches are separately trained by alternate learning. They are updated alternately with the guidance of each other. As a result, the learned similarity branch can better serve for the unseen compound FER task, given only a few reference images of novel classes.

\subsection{Joint Learning}

%\textcolor{red}{Intuitively, human can better recognize compound expressions with more knowledge of basic expressions. Therefore, }
%In the first stage, we jointly train the emotion branch and the similarity branch.
%This will {significantly} improve the inference ability of the similarity branch on basic expressions from the unseen domain, and thus facilitate the training of the second stage.
Different from conventional FSL benchmarks that contain a large number of classes,
basic expression datasets involve only a few basic expression {categories} (i.e., base classes in our setting). As a consequence, the constructed few-shot classification tasks are severely overlapped across different episodes, and existing FSL methods are likely to be trapped into the sampled base classes, leading to overfitting. To address this problem, we jointly train the emotion branch and the similarity branch. During the joint learning stage, the emotion branch, which captures the global information of basic expressions, is served as a regularizer to avoid overfitting of the similarity branch. Such a way  {significantly} improves the inference ability of the similarity branch on basic expressions from the unseen domain, and thus facilitates the training of the second stage. The optimization objective of this stage is formulated as
\begin{equation}
	\label{eq:joint}
	\mathcal{L}_{joint} = \mathcal{L}_{sim}+\lambda_{emo}\mathcal{L}_{emo},
\end{equation}
where $\mathcal{L}_{joint}$ denotes the joint loss. $\mathcal{L}_{sim}$ and $\mathcal{L}_{emo}$ represent the classification losses of the similarity branch and the emotion branch, respectively. $\lambda_{emo}$ denotes the balanced parameter.

In the following, we {will} introduce the emotion branch and the similarity branch in detail.

\noindent \textbf{Emotion Branch.}
The emotion branch consists of an emotion encoder $E_{e}$ and a {classifier} $f$ to classify the basic expressions. By performing the basic {FER task}, the emotion branch provides a global view of all basic expression information. %, which helps to alleviate the problem of highly-overlapped base classes in training the similarity branch.
Given multiple source domains $\mathbb{D}_{s}=\{D_{1},D_{2},$ $\cdots, D_{J} \}$, where $D_{j}$ represents the $j$-th source domain and $J$ is the {total} number of {training} domains, a source domain $D_{r}$ is randomly selected in every episode. The batch data $\{X_{i}^{r},Y_{i}^{r}\}$ are sampled from $D_{r}$, where $X_{i}^{r}$ and $Y_{i}^{r}$ denote the batch images and their corresponding labels, respectively.
The predicted label $\hat{y}_{i}^{r}$ is {then} computed as $\hat{y}_{i}^{r}=f(E_e(\bm{x}_{i}^{r}))$, where $\bm{x}_{i}^{r}$ denotes a single image from the sampled batch.
The classification loss of the emotion branch  $\mathcal{L}_{emo}$ employs the popular cross-entropy loss between the predicted result $\hat{y}_{i}^{r}$ and the ground-truth expression label $y_{i}^{r}$, that is,
\begin{equation}
	\label{eq:emo}
	\mathcal{L}_{emo} = -\sum_{c=1}^{C_r}\mathds{1}_{[c=y_i^{r}]}\log(f(E_e(\bm{x}_{i}^{r}))),
\end{equation}
where $C_r$ denotes the number of basic expression categories in $D_r$. {Indicator function} $\mathds{1}_{[c=y_i^{r}]}$ equals to 1 only if $c=y_i^{r}$, and 0 otherwise.

\noindent \textbf{Similarity Branch.}
%Following the traditional L2M settings,
The similarity branch involves a similarity encoder $E_{s}$ and a metric module $M$.
The similarity encoder $E_{s}$ and the emotion encoder $E_{e}$ share the parameters in the joint learning stage.
Mathematically, for a meta-training episode, given a randomly selected domain $D_{r}$, the training data are randomly sampled and divided into a support set $\mathbb{S}=\{X_{s}^{r}, Y_{s}^{r}\}$ and a query set $\mathbb{Q}=\{X_{q}^{r}, Y_{q}^{r}\}$, {where $X_{s}^{r}, Y_{s}^{r}$ and $X_{q}^{r}, Y_{q}^{r}$ }denote the sampled images and their corresponding labels in the support set and the query set, respectively.
Subsequently, an $N$-way $K$-shot classification task is constructed, where $N$ denotes the number of sampled classes and $K$ represents the number of the labeled images in each class of the support set.
The goal of a few-shot classification task is to make predictions for the query {images} with the reference of the support set.

All the images from the support set and the query set are fed into the similarity {branch} to evaluate the similarity between them. Then, the query image is assigned to its nearest category according to the similarity between this image and the support set in the learned feature space. The prediction process is formulated as

\begin{equation}
	\hat{Y_{q}^{r}} = g(M(E_{s}(X_{s}^{r}),E_s(X_{q}^{r})),Y_{s}^{r}),
\end{equation}
where $\hat{Y_{q}^{r}}$ represents the predicted results for the query images. $M(\cdot)$ denotes the metric function, and $g(\cdot)$ refers to the operation that assigns a query image to its nearest {category} according to the similarity metric. 

The objective of each few-shot classification task is to minimize the loss between the predicted result $\hat{y}_{q}^{r}$ and the ground-truth label $y_{q}^{r}$ of each query image as
\begin{equation}
	\label{eq:sim}
	\mathcal{L}_{sim} = -\sum_{n=1}^{N}\mathds{1}_{[n=y_q^{r}]}\log(\hat{y}_{q}^{r}).
\end{equation}
%where $\mathcal{L}_{sim}$ represents the classification loss of the similarity branch.

By training across different meta-tasks, the similarity branch can be easily adapted to an unseen task.

\subsection{Alternate Learning}
After joint learning, the inference ability of the similarity branch on basic expressions from the unseen domain is substantially improved, while that on compound expressions is still inferior. This is due to the poor inference ability of the initial emotion branch on novel classes.
Motivated by the observation that humans are able to better learn knowledge by communicating with each other from a different perspective, we further develop the alternate learning stage. This learning stage can be viewed as a ``two-student game'', where one student (branch) learns from the other one in turn.

More specifically, at the beginning of this stage, we update the emotion branch to perform its own expression classification task under the supervision of the fixed similarity branch for $K_e$ periods.
Given a sampled image $\bm{x}_{i}^{r}$, the objective function $\mathcal{L}_{emo}^{al}$ in this step is given as
\begin{equation}
	\mathcal{L}_{emo}^{al} = \mathcal{L}_{emo}+\theta_{n_e}||E_{s}(\bm{x}_{i}^{r}) -E_{e}(\bm{x}_{i}^{r})||^{2}_{2},
\end{equation}
where $\mathcal{L}_{emo}$ denotes the classification loss defined in Eq.~(\ref{eq:emo}) for the emotion branch. $\theta_{n_e}$ denotes the dynamic weight that varies with the episode $n_e$. In this paper, we adopt a weight decay strategy to highlight the key role of the emotion branch in this step during {alternate} learning. $||E_{s}(\bm{x}_{i}^{r}) -E_{e}(\bm{x}_{i}^{r})||_2$ is a regularized term, which constrains the feature distance between the similarity encoder and the emotion encoder to be as {close} as possible. Consequently, the emotion branch {captures} the knowledge that can be transferred to an unseen task to some extent.

Then, the role of each branch is exchanged, where the similarity branch intends to learn from the updated emotion branch in an episodic manner for $K_s$ periods. By resorting to the enhanced inference ability, the updated emotion branch can {boost} the classification performance of the similarity branch on both basic and compound expressions from the unseen domain. The objective function $\mathcal{L}_{sim}^{al}$ in this step is formulated as
\begin{equation}
	\mathcal{L}_{sim}^{al} = \mathcal{L}_{sim}+\theta_{n_e}||E_{e}(\bm{x}_{i}^{r}) -E_{s}(\bm{x}_{i}^{r})||^{2}_{2},
\end{equation}
where $\mathcal{L}_{sim}$ represents the {metric based} classification loss defined in Eq.~(\ref{eq:sim})  for the similarity branch. Similarly, we emphasize the {key} role of the similarity branch by the dynamic weight $\theta_{n_e}$  in this step.% during alternate learning.

Next, the similarity branch and the emotion branch are {alternately} trained several times. Unlike the ``two-player game'' of GAN \cite{goodfellow2014generative}, in which the generator and the discriminator compete with each other, the proposed alternate learning {stage} {improves} the inference ability of both branches by exchanging {their} respective knowledge. Finally, a similarity branch, which has superior inference ability on novel classes, can be obtained and transferred to perform the unseen compound FER task.

\subsection{Overall Training}
%The proposed two-stage learning framework to progressively train EGS-Net is summarized in Algorithm \ref{alg:2-stage}.
In the first stage, we jointly train 
the similarity branch and the emotion branch, preventing the similarity branch from overfitting to {highly overlapped} sampled {base} classes.
%The learned parameters of the two branches are then used as the initialized parameters for the subsequent training.
In the second stage,  we {alternately} train one branch with
the guidance of the other one. The similarity branch is first fixed while the emotion branch is updated to improve its inference ability.
%This is advantageous to facilitate the subsequent training of similarity branch.
Then, the similarity branch is optimized under the supervision of the updated emotion branch to {better} exploit the global information of basic expressions.
%Such a manner helps to regularize the training of similarity branch to avoid overfitting on sampled classes.
Finally, the above two steps are {alternately} trained. %\textcolor{red}{for $T_{al}$ epochs}.
{In this way, the two branches} can learn from each other from a different perspective, greatly improving the inference ability of {the} similarity branch to identify unseen compound expressions. {The two-stage learning framework is summarized in the appendix.}

\section{Experiments}

%In this section, we first describe the datasets and implementation details. Next, we perform extensive ablation studies to show the effectiveness of key components in our method and the superiority of the developed two-stage learning framework. Finally, we further analyze the results on different forms of similarity branches (various L2M baselines) to evaluate the extensibility of our method.

\subsection{Datasets}

In this paper, we study the compound FER task in the CD-FSL setting, where only  {images from easily accessible basic expression datasets} are used to train the model.
%To enrich the diversity of base classes and
%To bridge the domain gap between the training set and the test set,
We use several popular basic expression datasets, including three in-the-lab datasets (CK+ \cite{lucey2010extended}, MMI \cite{pantic2005web}, and Oulu-CASIA \cite{zhao2011facial}) and two in-the-wild datasets (RAF-DB \cite{li2017reliable} and SFEW \cite{dhall2011static}),  as multiple source domains for training. We use two newly released compound expression datasets (CFEE \cite{du2014compound} and EmotioNet \cite{benitez2017emotionet}) for testing. {More information of these datasets is provided in the appendix.}

To better analyze the inference ability of our method, we divide CFEE into two subsets, including 1,610 images labeled with basic expressions (denoted CFEE\_B) and 3,450 images labeled with compound expressions (denoted CFEE\_C).
%
%\noindent
%\textbf{EmotioNet}:
%EmotioNet is a large-scale in-the-wild database with one million facial expression images collected from the Internet.
%By using an AU-based annotation algorithm, most images are labeled automatically.
%In our experiment, we follow the second track of the EmotioNet Challenge \cite{benitez2017emotionet}, in which a dataset with six basic and ten compound expressions is used.
Similar to CFEE, we divide {EmotioNet} into EmotioNet\_B (consisting of basic expressions) and EmotioNet\_C (consisting of compound expressions).

\subsection{Implementation Details}

For all the experiments, we first align and crop facial images by MTCNN \cite{zhang2016joint}, and further resize them to 224 $\times$ 224.
All the images in basic expression datasets are used for training. The compound expression datasets and their corresponding subsets are used for testing.

We implement our model with the Pytorch toolbox. We adopt ResNet-18 \cite{he2016deep}  %\textcolor{red}{and additional task-specific layers are}
as the backbone for both the emotion encoder and the similarity encoder, which share the parameters in the joint learning stage and are separately updated in the alternate learning stage.
%The emotion encoder is pre-trained on the MS-Celeb-1M face recognition dataset \cite{guo2016ms}.
%We follow the public implementation for L2M methods\footnote{https://github.com/hytseng0509/CrossDomainFewShot}, and consider four classical FSL methods, including ProtoNet \cite{snell2017prototypical}, MatchingNet \cite{vinyals2016matching}, Relation Net \cite{sung2018learning}, and GNN Net \cite{garcia2017few}.
The networks are optimized by using the Adam algorithm \cite{kingma2014adam} with the learning rate of 0.001, $\beta_1=0.500$, and $\beta_2=0.999$. The weight of the emotion branch is empirically set to $\lambda_{emo}=1$ during the joint learning stage. We adopt the step decay strategy during the alternate learning stage.
For the emotion branch, the batch size is set to 128. For the similarity branch, we randomly sample $N~(=5)$ classes and $K~(=1, 5)$ images from each class to form the support set, and the number of query images is set to 16. %In ablation studies, we also evaluate the case when $K=1$.
The whole training contains %$T_{joint}=200$
200 epochs for joint learning and
%{$T_{al}=5$}
5 epochs for {alternate} learning, and the two branches exchange the role after every 20 {periods} ($K_e=K_s=20$). The number of episodes $N_e$  in each epoch is set to 100. We report the average recognition accuracy on 1000 meta-test tasks.

\begin{table*}[th!]
	\centering
	\resizebox{2.1\columnwidth}{!}{
		\begin{tabular}{l|c c|c c|c c|c c|c c|c c}
			\hline
			%\toprule [2 pt]
			\multirow{2}{*}{Method} & \multicolumn{2}{c|}{CFEE} & \multicolumn{2}{c|}{CFEE\_B} & \multicolumn{2}{c|}{CFEE\_C} & \multicolumn{2}{c|}{EmotioNet} & \multicolumn{2}{c|}{EmotioNet\_B} & \multicolumn{2}{c}{EmotioNet\_C} \\
			\cline{2-13}
			& 5-shot      & 1-shot      & 5-shot        & 1-shot       & 5-shot        & 1-shot       & 5-shot       & 1-shot        &
			5-shot        & 1-shot          & 5-shot          & 1-shot          \\
			\hline
			%\midrule[1 pt]
			E$\rm_b$ (single)             & 59.36       & 47.48           & 71.57         & 59.87        & 55.00         & 43.13        & 54.86          & 44.01             & 63.18           & 50.24           & 54.76           & 45.25           \\
			%\hline
			E$\rm_b$ (multiple)                    & 65.59       & 52.65       & 80.28         & 67.48           & 60.66         & 47.48            & 56.03          &  45.17        & 64.45           & 51.42           & 55.90           & 46.23               \\
			%\hline
			S$\rm_b$ (single)              & 65.41       & 54.22       & 71.44         & 65.73        & 62.69         & 50.07          & 56.35          & 46.38         & 63.13           & 51.12       & 57.66           & 48.13           \\
			%\hline
			S$\rm_b$ (multiple)                        & 69.69       & 58.05           & 82.21         & 72.51       & 66.84       & 54.30        & 57.49          &48.58         & 68.24       & 56.66           & 58.40           & 49.93           \\
			%\hline
			EGS-Net (joint)              & 70.88      & 59.18        & 85.63         & 76.74        & 67.05         & 54.99            & 58.57          & 49.14          & 70.60          & 59.16               & 58.83           & 50.57           \\
			%\hline
			EGS-Net (al)  &  71.25     & 60.02    &  84.09     &  75.11  & 67.33     & 55.13       &  58.73         & 49.28         & 69.39     &  57.32       & 59.25      & 50.86      \\
			%\hline
			EGS-Net            & \textbf{72.17}       & \textbf{60.90}   & \textbf{86.45}             & \textbf{77.16}       & \textbf{68.38}    & \textbf{56.65}       & \textbf{59.77}      & \textbf{50.06}    & \textbf{71.65}       & \textbf{59.67}    & \textbf{60.52}   & \textbf{51.62}     \\
			\hline
		\end{tabular}
	}
	\caption{The 5-shot and 1-shot accuracy (\%) on CFEE, EmotioNet, and the corresponding subsets.}
	\label{tab:all}
\end{table*}

\begin{table}[]
	\huge
	\resizebox{\columnwidth}{!}{
	%\begin{tabular}{p{2.035cm}|p{0.85cm}<{\centering}|p{1.2cm}<{\centering}|p{1.55cm}<{\centering}|p{1.92cm}<{\centering}}
		\begin{tabular}{l|c|c|c|c}
	
			\hline
			Method & CFEE & CFEE\_C & EmotioNet & EmotioNet\_C \\ \hline
			E$\rm_b$ (joint) & 69.14   & 65.46   & 57.61     & 57.44        \\ %\hline
			E$\rm_b$ (two-stage)  & \textbf{71.30} & \textbf{66.77} & \textbf{58.72}  & \textbf{59.25} \\
			%\bottomrule[1 pt]
			\hline
		\end{tabular}
	}
	\caption{Inference ability of the emotion branch. The 5-shot accuracy (\%) is reported for performance evaluation.}
	\label{tab:emo}
	
\end{table}

\begin{table}[]
	\huge
	\resizebox{1\columnwidth}{!}{
		%\begin{tabular}{p{2.035cm}<{\centering}|p{0.85cm}<{\centering}|p{1.2cm}<{\centering}|p{1.55cm}<{\centering}|p{1.92cm}<{\centering}}
			
		\begin{tabular}{p{3.52cm}<{\centering}|c|c|c|c}
		\hline
			  Weight decay & CFEE & CFEE\_C & EmotioNet & EmotioNet\_C \\ \hline
			$\times$    & 71.31  & 67.03   & 59.33    & 59.73       \\ %\hline
			\checkmark  & \textbf{72.17}    & \textbf{68.38}   & \textbf{59.77}    & \textbf{60.52}        \\ \hline
		\end{tabular}
	}
	\caption{Influence of the weight decay strategy. The 5-shot accuracy (\%) is reported for performance evaluation.}
	\label{tab:wd}
\end{table}

\subsection{Ablation Studies}
%We conduct ablation studies to demonstrate the importance of key components in EGS-Net and the effectiveness of our proposed two-stage learning framework.

To better analyze the inference ability on the unseen domain, we test the method on the whole dataset and two subsets, including a subset of basic expressions (CFEE\_B or EmotioNet\_B) and a subset of unseen compound expressions (CFEE\_C or EmotioNet\_C).
The whole dataset is used to evaluate the overall accuracy, while the subsets of basic and compound expressions are used to
evaluate the inference ability of a method on the seen classes and the novel classes from the unseen domain, respectively. The classical ProtoNet \cite{snell2017prototypical} is used as our similarity branch in this subsection.

\noindent \textbf{Influence of  Emotion Branch and Similarity Branch.} We first evaluate the inference ability of the emotion branch and the similarity branch,
when they are independently trained without using the two-stage learning framework. The two branches are evaluated in an FSL manner. That is, images from the support and query sets are fed into the trained feature extractor to extract features. The query images are then assigned to their nearest neighbors in the support set.
%, where a single domain (RAF-DB) and multiple source domains are respectively used for training.
We respectively denote the emotion branch and the similarity branch trained on a single domain (RAF-DB is used) as  {E$\rm_b$  (single) and S$\rm_b$  (single)}, and those trained on multiple source domains as  {E$\rm_b$ (multiple) and S$\rm_b$  (multiple)}. The comparison results are given in Table \ref{tab:all}.

As illustrated in Table \ref{tab:all}, S$\rm_b$ (single) and E$\rm_b$ (single) achieve similar performance for basic expression recognition on the unseen domains (i.e., CFEE\_B and EmotioNet\_B). In contrast, S$\rm_b$ (single) outperforms E$\rm_b$ (single) by a large margin for classifying unseen compound expressions (i.e., 7.69\%, 6.94\% on CFEE\_C, and 2.90\%, 2.88\% on EmotioNet\_C for 5-shot and 1-shot classification tasks, respectively).
{Similar patterns can be observed when multiple source domains are used.}
These results {indicate} that the inference ability of the similarity branch {on the unseen task} is better than that of the emotion branch. This can be ascribed to the superiority of the episodic training manner for the similarity branch.

%is because the similarity branch is trained across similar meta-tasks.
%Similar {patterns} can be observed when multiple source domains are used.
Moreover, E$\rm_b$ (multiple) and S$\rm_b$ (multiple) obtain much better recognition accuracy than E$\rm_b$ (single) and S$\rm_b$ (single), respectively, on the whole datasets and their corresponding subsets. %
%6.23\%, 4,28\% on CFEE, and 1.17\%, 1.14\% on EmotioNet for 5-shot classification task.
%The performance on their corresponding subsets are also enhanced.
Therefore, multiple source domains effectively enrich the diversity of the training data, and bridge the gap between the training set and the test set.
%In this way, the recognition accuracy obtained by  {E$\rm_b$ and S$\rm_b$} is greatly improved.
In the following part, we will use multiple source domains as the training set.

\begin{figure}[t!]
	\centering
	\includegraphics[scale=0.6]{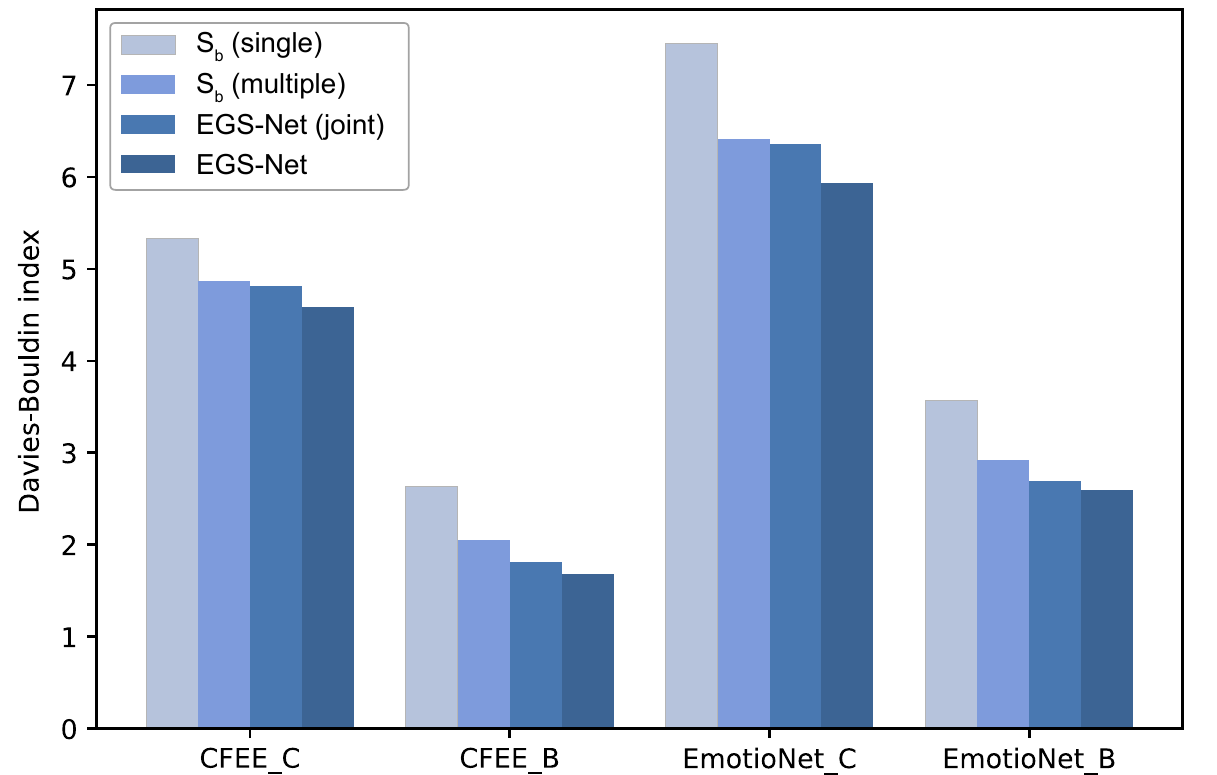}
	\caption{{DB index} of the learned features for basic and compound expressions on both in-the-lab and in-the-wild datasets. For {the DB index}, the smaller is better.}
	\label{fig:dbi}
\end{figure}

\begin{table*}[t]

	\centering
	\resizebox{1.77\columnwidth}{!}{
	\begin{tabular}{l|p{1.5cm}<{\centering}|p{1.5cm}<{\centering}|p{1.5cm}<{\centering}|p{1.5cm}<{\centering}}
		\hline
		
		\multirow{2}{*}{Method} & \multicolumn{2}{c|}{CFEE} & \multicolumn{2}{c}{EmotioNet}  \\
		\cline{2-5}
		& 5-shot      & 1-shot      & 5-shot        & 1-shot    \\
		\hline
		ProtoNet \cite{snell2017prototypical} & 69.69  & 58.05 & 57.49  & 48.58       \\ %\hline
		MatchingNet \cite{vinyals2016matching}& 64.70  & 56.75 & 54.14  & 48.09 \\
		RelationNet \cite{sung2018learning}& 65.27  & 56.51 & 56.18  & 48.33    \\ %\hline
		GNN \cite{garcia2017few}        & 70.10  & 58.45 & 58.06  & 49.23    \\
		%MAML \cite{finn2017model}    & 63.80  & 34.94  & 45.97  & 25.44   \\
		InfoPatch \cite{liu2021learning}        & 71.99  & 60.82 & 58.73  & 46.61  \\
		DKT \cite{patacchiola2020bayesian} & 67.55  & 54.94  & 55.30  & 45.39      \\
		GNN+LFT \cite{tseng2020cross} & 71.76  & 59.96   & 61.37  & 51.56    \\ %\hline
		\hline
		
		BASELINE \cite{chen2019closer} & 66.98  & 54.21 & 60.15  & 50.38    \\ %\hline
		BASELINE++ \cite{chen2019closer} & 68.60  & 56.28 & 61.13  & 51.00    \\ %\hline
		Arcmax loss \cite{afrasiyabi2020associative} & 68.92  & 56.94 & 60.87  & 51.02    \\ 	
		PT+NCM \cite{hu2020leveraging}     & 68.59 & 56.60 & 55.70  & 46.45     \\
		LR+DC \cite{yang2021free}     & 68.97 & 57.97 & 55.71  & 46.98 \\
		\hline
		EGS-Net (P) & 72.17  & 60.90 & 59.77  & 50.06     \\ %\hline
		EGS-Net (M) & 67.43  & 58.06 & 56.24  & 49.21    \\ %\hline
		EGS-Net (R) & 67.28  & 57.60  & 56.90  & 49.55   \\ %\hline
		EGS-Net (G) & \textbf{73.79}  & \textbf{61.28} & \textbf{62.12}  & \textbf{51.93}    \\ %\hline
		\hline
	\end{tabular}
	}
	\caption{The 5-shot and 1-shot accuracy (\%) comparisons among different competing methods on the in-the-lab CFEE and in-the-wild EmotioNet datasets.}
	\label{tab:sota}
	
\end{table*}

\noindent  \textbf{Influence of Joint Learning.}
The results obtained by EGS-Net only using the joint learning stage (denoted EGS-Net (joint)) are shown in Table  \ref{tab:all}. We also compare EGS-Net only using the alternate learning stage (denoted {EGS-Net (al)}) with EGS-Net using the two-stage learning framework.% \textcolor{red}{to demonstrate the significance of the progressive training manner}.

Compared with {S$\rm_b$ (multiple)}, EGS-Net (joint) obtains higher recognition accuracy on the basic expression subsets. To be specific, EGS-Net (joint) improves the performance by 3.42\%, 4.23\% on CFEE\_B, and 2.36\%, 2.50\%  on EmotioNet\_B for the 5-shot and 1-shot classification tasks, respectively.
Therefore, the joint learning stage is beneficial to alleviate the overfitting problem on sampled base classes, and {thus} enable EGS-Net to classify basic expressions from the unseen domain more accurately. In addition, compared with EGS-Net (al), EGS-Net respectively achieves the improvements of 1.05\% and 1.27\% in terms of recognition accuracy on CFEE\_C and EmotioNet\_C for the 5-shot classification task. This validates the necessity of the joint learning stage, which facilitates the training of the second stage.
%effectiveness to accomplish the unseen compound expression recognition task in the proposed progressive learning procedure.}

%Compared with S$\rm_b$ (multiple), the improvements obtained by EGS-Net (joint) on basic expression subsets (CFEE\_B and EmotioNet\_B) are larger than those on compound expression subsets (CFEE\_C and EmotioNet\_C). %In particular,EGS-Net (joint) gives a slight performance drop on CFEE\_C for the 5-shot classification task.
%This indicates that the inference ability of the similarity branch to generalize to the unseen task is still limited.

We also compute the Davies-Bouldin index (DB index) \cite{davies1979cluster} of different methods on four subsets. {DB index} depicts the intra-class variations and the inter-class similarities in the learned feature space.  {For {the DB index}, the smaller is better.} The results are shown in Figure \ref{fig:dbi}. We can observe that the {DB index} obtained by EGS-Net (joint) is {better} than that obtained by S$\rm_b$ (multiple) on all subsets. {However}, {the} DB index {decreases more} on basic expression {subsets} (i.e., CFEE\_B and EmotioNet\_B) than that on novel compound {subsets} (i.e., CFEE\_C and EmotioNet\_C). This shows that the inference ability of EGS-Net (joint) on unseen compound expressions is still inferior. The main reason is that the poor inference ability of the initial E$\rm_b$ (multiple) on the unseen task {constrains} the performance of {the} similarity branch during joint learning.

\noindent  \textbf{Influence of {Alternate} Learning.}
As shown in Table \ref{tab:all}, {EGS-Net (al)} gives higher recognition accuracy than $S\rm_b$ (multiple) on CFEE, EmotioNet, and their corresponding subsets. {Compared with EGS-Net (joint), EGS-Net (al) gives better accuracy on the compound subsets but performs worse on the basic subsets. This is because the alternate learning stage facilitates our model to identify unseen compound expressions by training the similarity branch across similar tasks. However, EGS-Net (al) still suffers from the overfitting problem caused by limited base classes.}

The two-stage EGS-Net further improves the performance of EGS-Net (joint),  especially for the unseen compound FER task.
Specifically, it obtains improvements of 1.33\%, 1.66\% on CFEE\_C, and 1.69\%, 1.05\% on EmotioNet\_C for the 5-shot and 1-shot classification tasks, respectively. Hence, the alternate learning stage is of great significance to enhance the inference ability of the similarity branch.

Moreover, we also evaluate the inference ability of the emotion branch, as shown in Table \ref{tab:emo}. We give the results obtained by the emotion branch based only on the joint learning stage (denoted E$\rm_b$ (joint)) and that based on the two-stage learning framework (denoted E$\rm_b$ (two-stage)).
We can see that the inference ability of the emotion branch on the unseen task is enhanced after the alternate learning stage {(1.31\% and 1.81\% improvements on CFEE\_C and EmotioNet\_C, respectively)}.
%under the guidance of the similarity branch.
Resorting to the improved inference ability, the emotion branch can better guide the training of the similarity branch in the second learning stage.

%\textcolor{red}
{In this paper, a weight decay strategy is introduced to highlight the key role of current training branch during the alternate learning stage. The influence of the weight decay strategy is shown in Table \ref{tab:wd}. We can observe that the weight decay strategy %used in the alternate learning stage 
is beneficial to improve the %final 
performance.} %(delete)

Finally, we demonstrate the discriminability of the learned features obtained by EGS-Net. From Figure \ref{fig:dbi}, EGS-Net gives {better} {DB index} than EGS-Net (joint) on four subsets. The gap is more evident on unseen compound expression subsets. This further validates the importance of the proposed {alternate} learning stage. {Furthermore, we also show some feature visualization results in the appendix.}
%Moreover, we visualize the learned feature space by t-SNE \cite{van2008visualizing} in Figure \ref{fig:fv_unseen}.  Compared with the learned features obtained by S$\rm_b$ (multiple), those obtained by EGS-Net are much more distinguishable on basic and compound subsets of CFEE.
%Thus, the proposed learning framework effectively enhances the inference ability of EGS-Net on both seen and unseen tasks from the unseen domain.

\subsection{Comparison with State-of-the-Art Methods}
\label{lab:analysis}

{Table \ref{tab:sota} gives the comparison results between our proposed method and several state-of-the-art FSL methods on the compound expression datasets. We build our EGS-Net methods based on four representative L2M methods, including ProtoNet \cite{snell2017prototypical}, MatchingNet \cite{vinyals2016matching}, RelationNet \cite{sung2018learning}, and GNN \cite{garcia2017few}, denoted EGS-Net (P), EGS-Net (M), EGS-Net (R), and EGS-Net (G), respectively. 
These methods differ in terms of metric modules. Specifically, ProtoNet and MatchingNet employ the Euclidean and cosine distances, respectively. A learnable metric module based on the vanilla and graph convolution is used in RelationNet and GNN.
For a fair comparison, all the competing methods are trained by
using publicly available codes under the same settings  (e.g., dataset and backbone).
%For a fair comparison, all the competing methods are implemented using ResNet-18 as the backbone.}

%These baselines differ mainly in the design of  as the metric me. %To be specific, ProtoNet uses the Euclidean distance while MatchingNet adopts the cosine distance to evaluate the similarity between the  and support images. A learnable module based on the vanilla and graph convolution is introduced in RelationNet and GNN as the metric function.

Compared with the corresponding L2M baselines, EGS-Net (P), EGS-Net (M), EGS-Net (R), and EGS-Net (G) achieve higher performance (2.48\%, 2,73\%, 2.01\%, 3.69\% improvements on CFEE, and 2.28\%, 2.10\%, 0.72\%, 4.06\% improvements on the more challenging EmotioNet dataset {for the 5-shot classification tasks}). The above results indicate {that} our proposed EGS-Net method {can} further improve the inference ability of existing L2M methods on the unseen compound expression datasets.

Moreover, we evaluate several recent FSL methods for performance comparison. 
For instance, InfoPatch \cite{liu2021learning} introduces contrastive learning into the episodic training manner for a general matching.
DKT \cite{patacchiola2020bayesian} learns a kernel that can transfer to a new task for the Bayesian model.
%InfoPatch \cite{liu2021learning} introduces contrastive learning into the episodic training, and DKT \cite{patacchiola2020bayesian} learns a transferable kernel for the Bayesian model.}  
Tseng \emph{et al.} \cite{tseng2020cross} solve the CD-FSL problem by using feature-wise transformation layers. 
Some transfer learning based methods focus on either the design of loss functions in the pretraining stage \cite{chen2019closer,afrasiyabi2020associative} or the calibration of novel class distribution in the fine-tuning stage \cite{hu2020leveraging,yang2021free}.
%\textcolor{red}{For a fair comparison, all the competing methods  are trained by using publicly available codes under the same settings as our method (e.g., dataset, backbone, and etc.).}
%The performance of these methods drops substantially in our FER task, where the number of base classes is limited. (delete)
As can be seen in Table \ref{tab:sota}, among all the competing methods, our EGS-Net (G), which uses a graph {convolution based} metric function, obtains the highest accuracy of 73.79\%, 61.28\% on the in-the-lab CFEE dataset, and 62.12\%, 51.93\% on the in-the-wild EmotioNet dataset for 5-shot and 1-shot classification tasks, respectively.

%This is mainly due to the use of a  predefined similarity function in ProtoNet
%ProtoNet uses  instead of the learnable metric module
%Compared with the learnable metric module, ProtoNet with a predefined similarity function This is mainly because the complexity of the model may also increase the overfitting problem of the training data.

%has a apparent impact on all the L2M baselines, which verifies that the limitation number of base classes influence the inference ability of original few-shot based methods severely. With the regularization of emotion branch in the joint training, the recognition accuracy achieve gains of \textcolor{blue}{xxx, xxx, xxx, xxx} respectively on the CFEE. It also improves the accuracy on EmotioNet dataset by \textcolor{blue}{xxx,xxx,xxx,xxx}. After training in a ``two-student game'' manner during the bidirectional learning process, the model performs better, \textcolor{blue}{especially for the unseen compound expressions.} Note that the improvement of the two-stage learning scheme on those learnable metric module is \textcolor{blue}{bigger} than that on the fixed metric function, and obtain the top accuracy on \textcolor{blue}{GNNNet of xxx}.

\section{Conclusion}
%In this paper, we propose a novel EGS-Net method, which is only trained on basic FER datasets, to classify novel compound expressions with very few reference images.

In this paper, we propose a novel EGS-Net method for compound FER in {the} CD-FSL setting, %By only training on basic expression datasets, EGS-Net can effectively classify unseen compound expressions with very few reference images
%Consequently, 
which substantially avoids the tedious collection of large-scale labeled compound expression training data and offers superior scalability for practical applications.
%for compound FER under the FSL paradigm. By only employing various basic FER benchmarks as our training set, our method
%
%Trained on various basic FER benchmarks, our method is  It
To alleviate the problem of limited base classes, a novel two-stage learning framework is developed. The proposed framework includes a joint learning stage to prevent the trained model from overfitting to highly overlapped sampled base classes, and an alternate learning stage to further improve the inference ability of our model for generalizing to the unseen task.
% is developed including
% is developed, including , and a bidirectional learning stage to which can be easily adapted to unseen task.
Extensive experiments have been performed to validate the effectiveness of our method on both in-the-lab and in-the-wild compound expression datasets.
%Further analysis on various L2M baselines has shown the great extensibility of {our proposed} EGS-Net.
%In future, we intend to evaluate our method on the more challenging fine-grained expression datasets.

\section{Acknowledgments}
This work was supported by the National Natural Science Foundation of China under Grants 62071404 and
61872307, by the Open Research Projects of Zhejiang Lab under Grant 2021KG0AB02, by the Natural Science Foundation of Fujian Province under Grant 2020J01001, and by the Youth Innovation Foundation of
Xiamen City under Grant 3502Z20206046.

\bibliography{ref}

\begin{thebibliography}{41}
\providecommand{\natexlab}[1]{#1}

\bibitem[{Afrasiyabi, Lalonde, and Gagn{\'e}(2020)}]{afrasiyabi2020associative}
Afrasiyabi, A.; Lalonde, J.-F.; and Gagn{\'e}, C. 2020.
\newblock Associative alignment for few-shot image classification.
\newblock In \emph{Proceedings of the European Conference on Computer Vision},
  18--35.

\bibitem[{Benitez-Quiroz et~al.(2017)Benitez-Quiroz, Srinivasan, Feng, Wang,
  and Martinez}]{benitez2017emotionet}
Benitez-Quiroz, C.~F.; Srinivasan, R.; Feng, Q.; Wang, Y.; and Martinez, A.~M.
  2017.
\newblock EmotioNet challenge: Recognition of facial expressions of emotion in
  the wild.
\newblock \emph{arXiv preprint arXiv:1703.01210}.

\bibitem[{Chen et~al.(2019)Chen, Liu, Kira, Wang, and Huang}]{chen2019closer}
Chen, W.-Y.; Liu, Y.-C.; Kira, Z.; Wang, Y.-C.; and Huang, J.-B. 2019.
\newblock A closer look at few-shot classification.
\newblock In \emph{Proceedings of the International Conference on Learning
  Representations}.

\bibitem[{Ciubotaru et~al.(2019)Ciubotaru, Devos, Bozorgtabar, Thiran, and
  Gabrani}]{ciubotaru2019revisiting}
Ciubotaru, A.-N.; Devos, A.; Bozorgtabar, B.; Thiran, J.-P.; and Gabrani, M.
  2019.
\newblock Revisiting few-shot learning for facial expression recognition.
\newblock \emph{arXiv preprint arXiv:1912.02751}.

\bibitem[{Corneanu et~al.(2016)Corneanu, Sim{\'o}n, Cohn, and
  Guerrero}]{corneanu2016survey}
Corneanu, C.~A.; Sim{\'o}n, M.~O.; Cohn, J.~F.; and Guerrero, S.~E. 2016.
\newblock Survey on RGB, 3D, thermal, and multimodal approaches for facial
  expression recognition: History, trends, and affect-related applications.
\newblock \emph{IEEE Transactions on Pattern Analysis and Machine
  Intelligence}, 38(8): 1548--1568.

\bibitem[{Davies and Bouldin(1979)}]{davies1979cluster}
Davies, D.~L.; and Bouldin, D.~W. 1979.
\newblock A cluster separation measure.
\newblock \emph{IEEE Transactions on Pattern Analysis and Machine
  Intelligence}, (2): 224--227.

\bibitem[{Dhall et~al.(2011)Dhall, Goecke, Lucey, and Gedeon}]{dhall2011static}
Dhall, A.; Goecke, R.; Lucey, S.; and Gedeon, T. 2011.
\newblock Static facial expression analysis in tough conditions: Data,
  evaluation protocol and benchmark.
\newblock In \emph{Proceedings of the IEEE International Conference on Computer
  Vision Workshops}, 2106--2112.

\bibitem[{Dong et~al.(2018)Dong, Zheng, Ma, Yang, and Meng}]{dong2018few}
Dong, X.; Zheng, L.; Ma, F.; Yang, Y.; and Meng, D. 2018.
\newblock Few-example object detection with model communication.
\newblock \emph{IEEE Transactions on Pattern Analysis and Machine
  Intelligence}, 41(7): 1641--1654.

\bibitem[{Du, Tao, and Martinez(2014)}]{du2014compound}
Du, S.; Tao, Y.; and Martinez, A.~M. 2014.
\newblock Compound facial expressions of emotion.
\newblock \emph{Proceedings of the National Academy of Sciences}, 111(15):
  E1454--E1462.

\bibitem[{Ekman and Friesen(1971)}]{ekman1971constants}
Ekman, P.; and Friesen, W.~V. 1971.
\newblock Constants across cultures in the face and emotion.
\newblock \emph{Journal of Personality and Social Psychology}, 17(2): 124--129.

\bibitem[{Fabian Benitez-Quiroz, Srinivasan, and
  Martinez(2016)}]{fabian2016emotionet}
Fabian Benitez-Quiroz, C.; Srinivasan, R.; and Martinez, A.~M. 2016.
\newblock EmotioNet: An accurate, real-time algorithm for the automatic
  annotation of a million facial expressions in the wild.
\newblock In \emph{Proceedings of the IEEE International Conference on Computer
  Vision and Pattern Recognition}, 5562--5570.

\bibitem[{Finn, Abbeel, and Levine(2017)}]{finn2017model}
Finn, C.; Abbeel, P.; and Levine, S. 2017.
\newblock Model-agnostic meta-learning for fast adaptation of deep networks.
\newblock In \emph{Proceedings of the International Conference on Machine
  Learning}, 1126--1135.

\bibitem[{Garcia and Bruna(2018)}]{garcia2017few}
Garcia, V.; and Bruna, J. 2018.
\newblock Few-shot learning with graph neural networks.
\newblock In \emph{Proceedings of the International Conference on Learning
  Representations}.

\bibitem[{Goodfellow et~al.(2014)Goodfellow, Pouget-Abadie, Mirza, Xu,
  Warde-Farley, Ozair, Courville, and Bengio}]{goodfellow2014generative}
Goodfellow, I.~J.; Pouget-Abadie, J.; Mirza, M.; Xu, B.; Warde-Farley, D.;
  Ozair, S.; Courville, A.; and Bengio, Y. 2014.
\newblock Generative adversarial networks.
\newblock \emph{arXiv preprint arXiv:1406.2661}.

\bibitem[{Guo et~al.(2017)Guo, Zhou, Wu, Wan, Zhu, Lei, and Li}]{guo2017multi}
Guo, J.; Zhou, S.; Wu, J.; Wan, J.; Zhu, X.; Lei, Z.; and Li, S.~Z. 2017.
\newblock Multi-modality network with visual and geometrical information for
  micro emotion recognition.
\newblock In \emph{Proceedings of the IEEE International Conference on
  Automatic Face and Gesture Recognition}, 814--819.

\bibitem[{Guo et~al.(2020)Guo, Codella, Karlinsky, Codella, Smith, Saenko,
  Rosing, and Feris}]{guo2020broader}
Guo, Y.; Codella, N.~C.; Karlinsky, L.; Codella, J.~V.; Smith, J.~R.; Saenko,
  K.; Rosing, T.; and Feris, R. 2020.
\newblock A broader study of cross-domain few-shot learning.
\newblock In \emph{Proceedings of the European Conference on Computer Vision},
  124--141.

\bibitem[{He et~al.(2016)He, Zhang, Ren, and Sun}]{he2016deep}
He, K.; Zhang, X.; Ren, S.; and Sun, J. 2016.
\newblock Deep residual learning for image recognition.
\newblock In \emph{Proceedings of the IEEE International Conference on Computer
  Vision and Pattern Recognition}, 770--778.

\bibitem[{Hu, Gripon, and Pateux(2021)}]{hu2020leveraging}
Hu, Y.; Gripon, V.; and Pateux, S. 2021.
\newblock Leveraging the feature distribution in transfer-based few-shot
  learning.
\newblock In \emph{Proceedings of the International Conference on Artificial
  Neural Networks}, 487--499.

\bibitem[{Kingma and Ba(2015)}]{kingma2014adam}
Kingma, D.~P.; and Ba, J. 2015.
\newblock Adam: A method for stochastic optimization.
\newblock In \emph{Proceedings of the International Conference on Learning
  Representations}.

\bibitem[{Lake, Salakhutdinov, and Tenenbaum(2015)}]{lake2015human}
Lake, B.~M.; Salakhutdinov, R.; and Tenenbaum, J.~B. 2015.
\newblock Human-level concept learning through probabilistic program induction.
\newblock \emph{Science}, 350(6266): 1332--1338.

\bibitem[{Li, Deng, and Du(2017)}]{li2017reliable}
Li, S.; Deng, W.; and Du, J. 2017.
\newblock Reliable crowdsourcing and deep locality-preserving learning for
  expression recognition in the wild.
\newblock In \emph{Proceedings of the IEEE International Conference on Computer
  Vision and Pattern Recognition}, 2852--2861.

\bibitem[{Li et~al.(2019)Li, Xu, Huo, Wang, Gao, and Luo}]{li2019distribution}
Li, W.; Xu, J.; Huo, J.; Wang, L.; Gao, Y.; and Luo, J. 2019.
\newblock Distribution consistency based covariance metric networks for
  few-shot learning.
\newblock In \emph{Proceedings of the AAAI Conference on Artificial
  Intelligence}, 8642--8649.

\bibitem[{Li et~al.(2018)Li, Zeng, Shan, and Chen}]{li2018occlusion}
Li, Y.; Zeng, J.; Shan, S.; and Chen, X. 2018.
\newblock Occlusion aware facial expression recognition using CNN with
  attention mechanism.
\newblock \emph{IEEE Transactions on Image Processing}, 28(5): 2439--2450.

\bibitem[{Liu et~al.(2021)Liu, Fu, Xu, Yang, Li, Wang, and
  Zhang}]{liu2021learning}
Liu, C.; Fu, Y.; Xu, C.; Yang, S.; Li, J.; Wang, C.; and Zhang, L. 2021.
\newblock Learning a few-shot embedding model with contrastive Learning.
\newblock In \emph{Proceedings of the AAAI Conference on Artificial
  Intelligence}, 8635--8643.

\bibitem[{Lu et~al.(2020)Lu, Gong, Ye, and Zhang}]{lu2020learning}
Lu, J.; Gong, P.; Ye, J.; and Zhang, C. 2020.
\newblock Learning from very few samples: A survey.
\newblock \emph{arXiv preprint arXiv:2009.02653}.

\bibitem[{Lucey et~al.(2010)Lucey, Cohn, Kanade, Saragih, Ambadar, and
  Matthews}]{lucey2010extended}
Lucey, P.; Cohn, J.~F.; Kanade, T.; Saragih, J.; Ambadar, Z.; and Matthews, I.
  2010.
\newblock The extended Cohn-Kanade dataset (CK+): A complete dataset for action
  unit and emotion-specified expression.
\newblock In \emph{Proceedings of the IEEE Computer Society Conference on
  Computer Vision and Pattern Recognition Workshops}, 94--101.

\bibitem[{Luo et~al.(2017)Luo, Zou, Hoffman, and Fei-Fei}]{luo2017label}
Luo, Z.; Zou, Y.; Hoffman, J.; and Fei-Fei, L. 2017.
\newblock Label efficient learning of transferable representations across
  domains and tasks.
\newblock In \emph{Advances in Neural Information Processing Systems},
  165--177.

\bibitem[{Pantic et~al.(2005)Pantic, Valstar, Rademaker, and
  Maat}]{pantic2005web}
Pantic, M.; Valstar, M.; Rademaker, R.; and Maat, L. 2005.
\newblock Web-based database for facial expression analysis.
\newblock In \emph{Proceedings of the IEEE International Conference on
  Multimedia and Expo}, 317--321.

\bibitem[{Patacchiola et~al.(2020)Patacchiola, Turner, Crowley, O'Boyle, and
  Storkey}]{patacchiola2020bayesian}
Patacchiola, M.; Turner, J.; Crowley, E.~J.; O'Boyle, M.; and Storkey, A. 2020.
\newblock Bayesian meta-learning for the few-shot setting via deep kernels.
\newblock In \emph{Advances in Neural Information Processing Systems}.

\bibitem[{Ruan et~al.(2020)Ruan, Yan, Chen, Xue, and Wang}]{ruan2020deep}
Ruan, D.; Yan, Y.; Chen, S.; Xue, J.; and Wang, H. 2020.
\newblock Deep disturbance-disentangled learning for facial expression
  recognition.
\newblock In \emph{Proceedings of the 28th ACM International Conference on
  Multimedia}, 2833--2841.

\bibitem[{Slimani et~al.(2019)Slimani, Lekdioui, Messoussi, and
  Touahni}]{slimani2019compound}
Slimani, K.; Lekdioui, K.; Messoussi, R.; and Touahni, R. 2019.
\newblock Compound facial expression recognition based on highway CNN.
\newblock In \emph{Proceedings of the New Challenges in Data Sciences: Acts of
  the Second Conference of the Moroccan Classification Society}, 1--7.

\bibitem[{Snell, Swersky, and Zemel(2017)}]{snell2017prototypical}
Snell, J.; Swersky, K.; and Zemel, R.~S. 2017.
\newblock Prototypical networks for few-shot learning.
\newblock In \emph{Advances in Neural Information Processing Systems},
  4077--4087.

\bibitem[{Sung et~al.(2018)Sung, Yang, Zhang, Xiang, Torr, and
  Hospedales}]{sung2018learning}
Sung, F.; Yang, Y.; Zhang, L.; Xiang, T.; Torr, P.~H.; and Hospedales, T.~M.
  2018.
\newblock Learning to compare: Relation network for few-shot learning.
\newblock In \emph{Proceedings of the IEEE International Conference on Computer
  Vision and Pattern Recognition}, 1199--1208.

\bibitem[{Tseng et~al.(2020)Tseng, Lee, Huang, and Yang}]{tseng2020cross}
Tseng, H.-Y.; Lee, H.-Y.; Huang, J.-B.; and Yang, M.-H. 2020.
\newblock Cross-domain few-shot classification via learned feature-wise
  transformation.
\newblock In \emph{Proceedings of the International Conference on Learning
  Representations}.

\bibitem[{Vinyals et~al.(2016)Vinyals, Blundell, Lillicrap, Kavukcuoglu, and
  Wierstra}]{vinyals2016matching}
Vinyals, O.; Blundell, C.; Lillicrap, T.; Kavukcuoglu, K.; and Wierstra, D.
  2016.
\newblock Matching networks for one shot learning.
\newblock In \emph{Advances in Neural Information Processing Systems},
  3630--3638.

\bibitem[{Yang, Liu, and Xu(2021)}]{yang2021free}
Yang, S.; Liu, L.; and Xu, M. 2021.
\newblock Free lunch for few-shot learning: Distribution calibration.
\newblock In \emph{Proceedings of the International Conference on Learning
  Representations}.

\bibitem[{Yang et~al.(2020)Yang, Wang, Chen, Liu, and Qiao}]{yang2020context}
Yang, Z.; Wang, Y.; Chen, X.; Liu, J.; and Qiao, Y. 2020.
\newblock Context-transformer: Tackling object confusion for few-shot
  detection.
\newblock In \emph{Proceedings of the AAAI Conference on Artificial
  Intelligence}, 12653--12660.

\bibitem[{Yao et~al.(2020)Yao, Zhang, Wei, Jiang, Wang, Huang, Chawla, and
  Li}]{yao2020graph}
Yao, H.; Zhang, C.; Wei, Y.; Jiang, M.; Wang, S.; Huang, J.; Chawla, N.; and
  Li, Z. 2020.
\newblock Graph few-shot learning via knowledge transfer.
\newblock In \emph{Proceedings of the AAAI Conference on Artificial
  Intelligence}, 6656--6663.

\bibitem[{Zhang et~al.(2016)Zhang, Zhang, Li, and Qiao}]{zhang2016joint}
Zhang, K.; Zhang, Z.; Li, Z.; and Qiao, Y. 2016.
\newblock Joint face detection and alignment using multitask cascaded
  convolutional networks.
\newblock \emph{IEEE Signal Processing Letters}, 23(10): 1499--1503.

\bibitem[{Zhao et~al.(2011)Zhao, Huang, Taini, Li, and
  Pietik{\"a}Inen}]{zhao2011facial}
Zhao, G.; Huang, X.; Taini, M.; Li, S.~Z.; and Pietik{\"a}Inen, M. 2011.
\newblock Facial expression recognition from near-infrared videos.
\newblock \emph{Image and Vision Computing}, 29(9): 607--619.

\bibitem[{Zhao, Liu, and Zhou(2021)}]{zhao2021robust}
Zhao, Z.; Liu, Q.; and Zhou, F. 2021.
\newblock Robust lightweight facial expression recognition network with label
  distribution training.
\newblock In \emph{Proceedings of the AAAI Conference on Artificial
  Intelligence}, 3510--3519.

\end{thebibliography}

\end{document}